\documentclass[12pt]{article}

\usepackage{amsmath} 
\usepackage{amsthm}

\usepackage[T1]{fontenc}
\usepackage{newtxtext}
\usepackage{newtxmath}
\usepackage{bm}

\usepackage[english]{babel}

\usepackage[parfill]{parskip}
\usepackage{afterpage}
\usepackage{framed}

\usepackage{enumitem}

\usepackage{setspace}

\usepackage[defaultlines=4,all]{nowidow}
\noclub[3]

\RequirePackage[bf,small]{titlesec}

\usepackage[usenames,dvipsnames]{xcolor}
\definecolor{shadecolor}{gray}{0.9}
\definecolor{shadecolor}{gray}{0.9}

\usepackage{graphicx}
\usepackage[labelfont=bf]{caption}
\usepackage[format=hang]{subcaption}

\usepackage[algoruled]{algorithm2e}
\usepackage{listings}
\usepackage{fancyvrb}
\fvset{fontsize=\normalsize}

\lstdefinestyle{mystyle}{
    commentstyle=\color{OliveGreen},
    keywordstyle=\color{BurntOrange},
    numberstyle=\tiny\color{black!60},
    stringstyle=\color{MidnightBlue},
    basicstyle=\ttfamily,
    breakatwhitespace=false,
    breaklines=true,
    captionpos=b,
    keepspaces=true,
    numbers=left,
    numbersep=5pt,
    showspaces=false,
    showstringspaces=false,
    showtabs=false,
    tabsize=2
}
\usepackage[
  paper  = letterpaper,
  left   = 1.25in,
  right  = 1.25in,
  top    = 1.0in,
  bottom = 1.0in,
  ]{geometry}

\usepackage{natbib}
\usepackage[colorlinks,linktoc=all]{hyperref}
\usepackage[all]{hypcap}
\hypersetup{citecolor=MidnightBlue}
\hypersetup{urlcolor=MidnightBlue}
\hypersetup{linkcolor=Black}

\usepackage[nameinlink]{cleveref}
\creflabelformat{equation}{#1#2#3}

\usepackage
[acronym,smallcaps,nowarn,section,nogroupskip,nonumberlist]{glossaries}
\glsdisablehyper{}
\setglossarystyle{long3col}
\makeglossaries

\newacronym{IMI}{imi}{instantaneous mutual information}
\newacronym{LDA}{lda}{latent Dirichlet allocation}
\newacronym{PPC}{ppc}{posterior predictive check}
\newacronym{PoPC}{pop-pc}{population predictive check}
\newacronym{DP}{dp}{Dirichlet process}

\usepackage{authblk}
\title{Accuracy on In-Domain Samples Matters When Building Out-of-Domain detectors:\\A Reply to Marek et al. (2021)}
\author[1]{Yinhe Zheng \thanks{zhengyinhe1@163.com}}
\author[2]{Guanyi Chen \thanks{g.chen@uu.nl}}
\affil[1]{Tsinghua University and Lingxin AI}
\affil[2]{Utrecht University}
\date{}

\begin{document}

\maketitle

\begin{abstract}
We have noticed that \citet{marek2021oodgan} try to re-implement our paper \citet{zheng2020out} in their work ``OodGAN: Generative Adversarial Network for Out-of-Domain Data Generation''. Our paper proposes a model to generate pseudo OOD samples that are akin to IN-Domain (IND) input utterances. These pseudo OOD samples can be used to improve the OOD detection performance by optimizing an entropy regularization term when building the IND classifier. \citet{marek2021oodgan} report a large gap between their re-implemented results and ours on the CLINC150 dataset \citep{larson2019evaluation}. This paper discusses some key observations that may have led to such a large gap. Most of these observations originate from our experiments because \citet{marek2021oodgan} have not released their codes\footnote{We have contacted the authors of \citet{marek2021oodgan}. However, they reply that all codes they wrote for their project are part of their internal codebase and hence unfortunately, cannot be shared.}. One of the most important observations is that stronger IND classifiers usually exhibit a more robust ability to detect OOD samples. We hope these observations help other researchers, including \citet{marek2021oodgan}, to develop better OOD detectors in their applications. \footnote{Our source code is at \url{https://github.com/silverriver/OOD4NLU}.}
\end{abstract}

We first recap the method proposed in \citet{marek2021oodgan} for generating pseudo OOD samples and training OOD detectors. Then we discuss some observations from our experiments of why a low performance was obtained by the implementation of \citet{marek2021oodgan}.

\section{Recap \citet{marek2021oodgan}}

\subsection{Generating Pseudo OOD Samples}

Detecting Out-of-Domain (OOD) inputs is important for NLU modules. However, it is challenging to collect high-quality OOD samples before training the OOD detector. Therefore, we proposed a pseudo OOD sample generation (POG) method that can produce pseudo OOD samples based on In-Domain (IND) inputs \citep{zheng2020out}. Specifically, we first build a continuous latent space for IND samples using an autoencoder and then introduce a generative adversarial network (i.e., a generator and a discriminator) to mimic that continuous latent space. An auxiliary classifier is further introduced to regularize the generated OOD samples to have indistinguishable intent labels.

Based on our approach, \citet{marek2021oodgan} propose to operate on the discrete space of token sequences directly. Specifically, their model also consists of a generator, a discriminator, and an auxiliary classifier. The major difference is that they directly employ the SeqGAN \citep{yu2017seqgan} approach to optimize their model.

\subsection{Training OOD detectors}

After obtaining pseudo OOD samples, \citet{marek2021oodgan} follows our work to train an IND classifier $P_\theta$ by optimizing the entropy regularization loss $\mathcal{L}_{ce}$ on pseudo OOD samples $\mathcal{D}_{ood}$ and the cross entropy loss $\mathcal{L}_{ent}$ on IND samples $\mathcal{D}_{ind}$:
\begin{equation}\label{eq:loss}
    \mathcal{L} 
    = \mathcal{L}_{ce} + \mathcal{L}_{ent} 
    = \mathop{\mathbb{E}}_{(x_i,y_i) \sim \mathcal{D}_{ind}}[-log P_{\theta}(y=y_i|x_i)] 
    +\mathop{\mathbb{E}}_{\hat{x} \sim \mathcal{P}_{ood}}[- \mathcal{H}(P_{\theta}(\bm{y}|\hat{x}))],
\end{equation}
in which $\theta$ is the model weights and $\mathcal{H}$ is the Shannon entropy of the predicted distribution. Assume there are total $K$ classes in IND samples, then $P_\theta$ is a $K$ classifier. The softmax output of $P_\theta$ is used to determine whether an input $x$ is an OOD sample or not. Specifically, the maximum score in the softmax output is computed:
\begin{equation}\label{eq:ood_score}
    Score(x) = \max_{i\in \{1,2,...,K\} } P_{\theta}(y=l_i|x),
\end{equation}
and $x$ is determined to be an OOD sample if $Score(x)$ is less than a threshold $\eta$. Note that $P_\theta$ can be used to classify IND samples and detect OOD samples simultaneously.

\section{Discussion}

\citet{marek2021oodgan} reported their re-implemented results of our model on the CLINC150 dataset \citep{larson2019evaluation}. However, their re-implemented results are much lower than ours. Specifically, they report an AUROC of 88.79, AUPR of 58.22, FPR95 of 36.49, FPR90 of 26.87. These metrics are much higher in our experiments: AUROC of 95.4, AUPR of 98.9, FPR95 of 25.0, FPR90 of 10.1. Here, we provide several hypothesizes of why such a large performance gap happens.

\subsection{Accuracy on IND Samples Matters}

We suspect the primary reason for such a performance gap is that \citeauthor{marek2021oodgan} did not train a good classifier for IND samples when optimizing Eq.\ref{eq:loss}. In fact, the OOD detection performance of $P_\theta$ largely depends on the accuracy of $P_\theta$ on IND samples. Generally, if $P_\theta$ performs well on classifying IND samples, it usually captures good features. Thereby it can perform well on OOD detection. Similar observations are also reported by \citet{hendrycks-etal-2020-pretrained}.

\begin{figure}[t]
  \centering
  \includegraphics[width=250px]{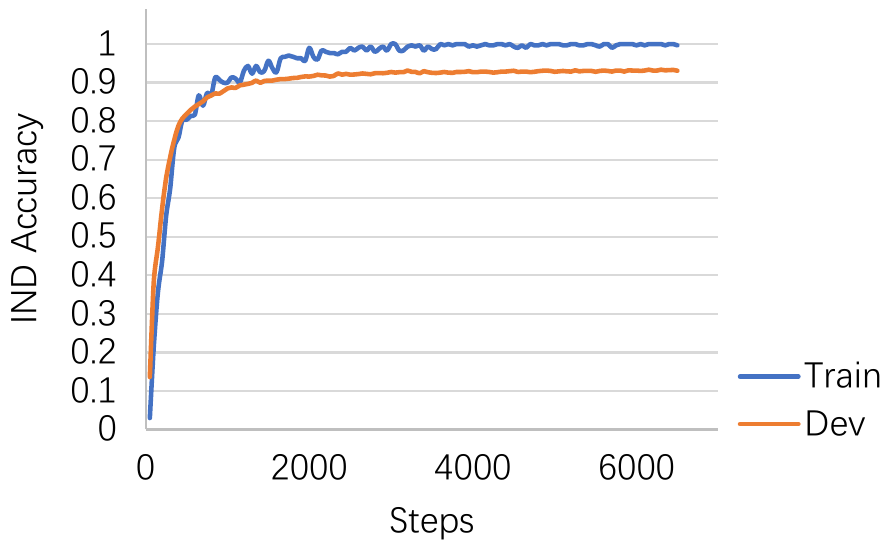}
  \caption{Train and Dev Accuracy for IND samples on the CLINC150 dataset using a CNN-based text classifier.}
  \label{fig:ind_acc}
\end{figure}

Specifically, the re-implemented model of \citeauthor{marek2021oodgan} obtains a rather low IND accuracy score of 88.00\% on the CLINC150 dataset. However, a simple CNN-based text classifier can push this score to 93.0+\% (see Figure \ref{fig:ind_acc}), and a BERT-based classifier \citep{devlin2018bert} reaches 97.00\%. We suspect such a degenerated classifier is the main reason for the low OOD detection performance observed by \citeauthor{marek2021oodgan}.

Note that optimizing the entropy regularization term $\mathcal{L}_{ent}$ on OOD samples has very little impact on the final classifier's IND accuracy. Similar results are reported in various previous studies \citep{hendrycks2018deep,lee2017training}.

\subsection{Pseudo OOD samples should not overlap with IND samples}

Another possible reason for the low performance reported by \citet{marek2021oodgan} is that they may have generated too many pseudo OOD samples that are semantically similar to IND examples. One possible reason is that the signal provided by the auxiliary classifier is not properly used to guide the generator. Specifically, \citet{marek2021oodgan} first train an auxiliary classifier and then fix it during the adversarial training. However, in the early training stage, the auxiliary classifier is too strong to guide the generator. Thus, the model may decide to ignore signals from the auxiliary classifier in the following steps. This is a common issue in training adversarial networks \citep{NIPS2014_5ca3e9b1}.

Note that similar issues are also observed in our early experiments. We tackled this issue by optimizing the auxiliary classifier from scratch together with our adversarial networks. Our approach effectively alleviates the issue of generating too many pseudo OOD samples that carry IND intents. Moreover, careful post-processing of the generated pseudo OOD samples also helps to alleviate this issue.

\subsection{Hyper-parameters Matters}

\citeauthor{marek2021oodgan} reported some of their model design and hyper-parameter settings used in their experiments, but many of them are not appear to be optimal.

Specifically, \citeauthor{marek2021oodgan} reported that they used fastText embeddings trained on Wikipedia for the generator. 
Though it is not fully clear how large their pre-training corpus is, it has been suggested that the GloVe embeddings that are trained on larger corpora than Wikipedia \citep{zheng2019persona,zheng2019personalized,zhuang2018quantifying} is a better option. 
Moreover, as the pre-training based generative model is becoming the de facto standard for text generation tasks \citep{zheng2020pre,zhang2020dialogue,zheng2021stylized,wang2020large,wang2021diversifying,wu2021transferable,he2021galaxy,zheng2021mmchat,zhou2021eva,liu2021dialoguecse,he2022unified}, replacing the generator with a pre-trained GPT model \citep{radfordimproving} would be a promising direction to pursue. 

\subsection{Comparing with More Baselines}
We also observe that the results of \citet{marek2021oodgan} on the CLINC150 dataset under-perform the simplest baseline: train a text classifier only on IND samples and use the maximum softmax score in Eq.\ref{eq:ood_score} to determine OOD samples. It needs no extra OOD samples.

In our experiments on CLINC150, such a simple baseline yields AUROC of 92.86, AUPR of 98.24, FPR95 of 39.72, FPR90 of 21.76. Similar results for this simple baseline are reported in other papers, including~\citet{podolskiy2021revisiting,yilmazd2u,ouyang-etal-2021-energy,wang2022practical,khosla-gangadharaiah-2022-evaluating}. In most above papers, its performance can reach an AUROC score of about 93.0, much higher than the method of \citeauthor{marek2021oodgan}.

\section{Conclusion}

In this paper, we discuss some key points that may lead to a poor OOD detector.
We hope our discussions benefit our readers in building better OOD detection models. 

\bibliographystyle{apalike}
\bibliography{bib.bib}

\end{document}